\documentclass[10pt,oneside,final]{IEEEtran}
\usepackage{amssymb}
\usepackage{amsmath}

\usepackage{graphicx}
\usepackage{color}
\usepackage{multicol}

\usepackage{subfigure}
\usepackage{bm}
\usepackage{latexsym}
\usepackage{stfloats}
\usepackage{float}
\usepackage{exscale}
\usepackage{relsize}
\usepackage{cite}
\usepackage{cases}
\usepackage{algorithm}
\usepackage{algpseudocode}
\usepackage{amsthm}
\usepackage{epsfig}
\usepackage{epstopdf}
\usepackage{breqn}
\usepackage{enumerate}
\usepackage{multirow}
\usepackage[utf8]{inputenc}
\usepackage{algorithmicx}
\usepackage{amssymb}
\usepackage{makecell}
\begin{document}
	\title{ WiFi-based Cross-Domain Gesture Recognition \\
		Using Attention Mechanism}
	\author{Ruijing Liu, 
		Cunhua Pan, ~\IEEEmembership{Senior Member,~IEEE, }
		Jiaming Zeng, 
		Hong Ren, ~\IEEEmembership{Member,~IEEE, }\\
		Kezhi Wang, ~\IEEEmembership{Senior Member,~IEEE, }
		Lei Kong,
		Jiangzhou Wang, ~\IEEEmembership{Fellow,~IEEE }
		\thanks{Ruijing Liu, Cunhua Pan, jiaming Zeng, Hong Ren are with the National Mobile Communications Research Laboratory, Southeast University, Nanjing 210096, China
			(e-mail: 230258158@seu.edu.cn; 
			cpan@seu.edu.cn;
			220250992@seu.edu.cn;	
			hren@seu.edu.cn).
			
			Kezhi Wang is with the Department of Computer Science, Brunel University
			London, UB8 3PH Uxbridge, U.K. (e-mail: kezhi.wang@brunel.ac.uk).
			
			Lei Kong is with the New H3C Technologies Co., Ltd $\&$ Zhejiang
			University (e-mail: kong.lei@h3c.com).
			
			Jiangzhou Wang is with the National Mobile Communications Research Laboratory, Southeast University, Nanjing 210096,
			China, and also with the Pervasive Communication Research Center, Purple
			Mountain Laboratories, Nanjing 211111, China (e-mail: j.z.wang@seu.edu.cn).}
	}
	\maketitle
	\newtheorem{lemma}{Lemma}
	\newtheorem{theorem}{Theorem}
	\newtheorem{remark}{Remark}
	\newtheorem{corollary}{Corollary}
	\newtheorem{proposition}{Proposition}
	\begin{abstract} 
     While fulfilling communication tasks, wireless signals can also be used to sense the environment. Among various types of sensing media, WiFi signals offer advantages such as widespread availability, low hardware cost, and strong robustness to environmental conditions like light, temperature, and humidity.  By analyzing Wi-Fi signals in the environment, it is possible to capture dynamic changes of the human body and accomplish sensing applications such as gesture recognition. Although many existing gesture sensing solutions perform well in-domain but lack cross-domain capabilities (i.e., recognition performance in untrained environments). To address this, we extract Doppler spectra from the channel state information (CSI) received by all receivers and concatenate each Doppler spectrum along the same time axis to generate fused images with multi-angle information as input features.
    Furthermore, inspired by the convolutional block attention module (CBAM), we propose a gesture recognition network that integrates a multi-semantic spatial attention mechanism with a self-attention-based channel mechanism. This network constructs attention maps to quantify the spatiotemporal features of gestures in images, enabling the extraction of key domain-independent features. Additionally, ResNet18 is employed as the backbone network to further capture deep-level features. To validate the network performance, we evaluate the proposed network on the public Widar3 dataset, and the results show that it not only maintains high in-domain accuracy of 99.72\%, but also achieves high performance in cross-domain recognition of 97.61\%, significantly outperforming existing best solutions.
	\end{abstract}
	\section{Introduction}

With the provision of emerging services such as the Internet of things (IoT) and smart homes, which demand high communication and sensing capabilities, integrated sensing and communication (ISAC) has been recognized as a key technology for the sixth generation (6G) wireless networks \cite{1}. Traditional wireless networks are also evolving from pure communication networks to joint communication and radar/radio sensing systems. By extracting environmental target information from received wireless signals, their functions extend beyond the conventional radar capabilities like localization, tracking, and motion recognition \cite{2}. Existing research on wireless sensing has already explored applications such as using radio frequency identification (RFID) technology for locating objects inside buildings \cite{3}, and utilizing received downlink signals for target localization within a practical cooperative ISAC network compatible with 5G NR standards \cite{4}.

Since human perception of the physical world has entered a new stage of ubiquitous intelligence, using gestures to convey instructions to computers offers a simple and direct approach for sending commands to smart devices such as those in smart homes. Traditional gesture recognition typically rely on wearable devices \cite{5} or cameras \cite{6} for capturing feature. However, such sensing approaches are often inconvenient or impose significant requirements on the sensing environment, such as lighting conditions. Additionally, these intrusive sensing methods raise concerns regarding security and privacy. Since WiFi equipment is widely deployed, utilizing WiFi signals for sensing can significantly reduce deployment costs while offering strong practicality and generalization in real-world applications. Moreover, utilizing WiFi signals for sensing eliminates the need to install surveillance cameras or various contact-based sensors in homes, enabling non-intrusive perception tasks \cite{7}. By analyzing WiFi signals in the environment, we can accurately capture dynamic changes in human and object activities, supporting a wide range of wireless sensing applications such as motion detection for human/nonhuman subjects \cite{8}, sleep monitoring \cite{9}, and fall detection \cite{10}.

Human motion affects the propagation of WiFi signals (e.g., through reflection, diffraction, and scattering), which enables the capture of human motion by analyzing received signals \cite{11}. WiFi-based sensing technologies can utilize channel state information (CSI) or received signal strength indicator (RSSI) to characterize wireless channel feature. Capturing the rising and falling edges in RSSI caused by hand movements relative to the device, different gestures can be distinguished \cite{12}. However, RSSI measures the combined effect of multi-path signal propagation, whereas CSI simultaneously measures the frequency response of multiple subcarriers within a single data packet rather than the overall amplitude response of all subcarriers combined. As a result, CSI is more sensitive and stable for such applications.

Methods for obtaining sensing information from CSI can be broadly categorized into two types: one employs mathematical modeling to reconstruct the relationship between CSI perturbation and gesture motions, while the other uses deep learning (DL) to extract high-level features of gestures from CSI. However, gesture recognition based on mathematical modeling imposes strict requirements on user positions, and several gestures cannot be distinguished across all possible locations \cite{13}. With advancements in DL, integrating communication technologies with DL has demonstrated potential to enhance system performance \cite{14}. Since DL can overcome the limitations of modeling-based methods and unlock more powerful and intelligent sensing capabilities, Widar3 \cite{15} employed a spatiotemporal DL neural network (DNN) to extract gesture-embedded spatiotemporal features, but the proposed model failed to capture deep-level feature information and did not fully exploit the potential of DL networks. FewSense \cite{16} adopted a few-shot learning framework, achieving excellent in-domain performance with minimal samples, but its cross-domain generalization remains limited due to insufficient adaptability . To better capture global features, the authors of \cite{17} proposed combining self-attention and relation-attention modules to extract gesture features, but this requires processing CSI into two distinct feature inputs for cooperative network training. Therefore, there is a need for a lightweight and streamlined DL network for gesture recognition that eliminates cumbersome preprocessing steps while effectively extracting deep features to improve cross-domain recognition accuracy across diverse scenarios.

To solve this, we propose an attention mechanism-based gesture recognition system that incorporates a shareable multi-semantic spatial attention (SMSA) \cite{18} and a self-attention-based channel attention. The attention mechanism demonstrates excellent learning effectiveness for time series, which can effectively extract domain-independent features to enhance the cross-domain recognition accuracy. By leveraging computer vision principles, we enable computers to interpret images through simulated human visual perception, facilitating advanced analysis and understanding. Specifically, we first preprocess CSI data using the conjugate multiplication method \cite{19} to eliminate random phase offsets caused by the lack of synchronization in commercial WiFi transceivers. Then we apply the short-time fourier transform (STFT) to extract Doppler frequency shift (DFS) information, and fuse the Doppler spectrograms of all receivers into standardized time-series images. This allows us to leverage visual perception techniques for training DL networks. Based on this, we design a gesture recognition network that jointly employs SMSA and self-attention-based channel attention mechanism. By constructing attention maps to quantify key spatiotemporal features of gestures within images, we integrate attention weights with corresponding pixel data to enable intelligent processing of DFS-fused images. Additionally, we adopt ResNet18 as the backbone network, leveraging its ability to facilitate feature reuse (preserving information integrity) and capture deep-level features.

The main contributions are summarized as follows:
\begin{itemize}
\item[$\bullet$]We investigate an attention mechanism-based gesture recognition system in WiFi sensing. The network integrates a SMSA module, which significantly enhances the comprehension of deep-level features, and further adopt a self-attention-based channel attention module to improve the capture of global feature information. By constructing attention maps to quantify spatiotemporal feature information of gestures in images, the network can effectively achieve the extraction of key feature information related to domain independence. ResNet18 is adopted as the backbone network
due to its ability to facilitate feature reuse to preserve
information integrity and capture deep-level features.
\item[$\bullet$]We propose an effective feature visualization method, which employs the conjugate multiplication method to eliminate random phase offsets caused by transmitter-receiver asynchronization, and further fuses DFS information extracted from CSI data collected by distributed receivers into time-series images to achieve fine characterization of gesture features. It not only can effectively utilize computer vision technology, but also enable intuitive identification of incomplete or defective data within the dataset.
\item[$\bullet$]We evaluate the proposed model on the public Widar3 dataset, and the proposed model achieves an in-domain average recognition accuracy of 99.72{\%} and a cross-domain average accuracy of 97.61{\%}, significantly outperforming existing researches.
\end{itemize}

The structure of this paper is organized as follows: Section II systematically reviews representative prior studies in the field of gesture recognition employing diverse signal processing methods for extracting CSI from WiFi signals. Section III elaborates on the joint attention network constructed based on the ResNet backbone architecture. Section IV provides the evaluation of experimental results base on the Widar3 dataset. Finally, Section V concludes the paper with a comprehensive summary.
\begin{figure}[t]
	\centering
	\includegraphics[width=0.9\linewidth]{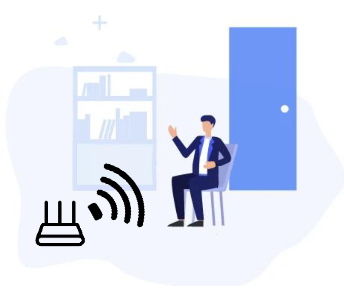}
	\caption{{WiFi Sensing Gesture Recognition.} } 
	\label{fig1}
\end{figure}

\section{Related Work} 

As shown in Fig. \ref{fig1}, the WiFi signals transmitted by device deployed indoor will contain gesture feature information. To extract the gesture feature from Wifi signal, the input features after preprocessing for networks can be categorized into three types: motion-based features, CSI-based features, and DFS-based features. For ease of reference, we have systematically summarized the representative studies listed below in Table \ref{table6}.

\begin{table*}[ht]
	\centering
	\caption{Related Work} 
	\label{table6}
	\scalebox{0.9}{
		\renewcommand{\arraystretch}{1.2}
		\begin{tabular}{|c|c|c|c|c|}
			\hline  
			Network & Feature Type & Method & Dataset & Performance \\
			\hline  
			Widar3 \cite{15} & BVP & CNN+GRU  & Widar3 & Gesture, IN-domain 92.7\%, Corss-domain 92.4\% \\
			\hline
			WiHF \cite{20} &	Motion variation pattern&	CNN+GRU &	Widar3 &	Gesture, IN-domain 97.65\%, Corss-domain 91.07\% \\
			\hline
			AutoFi \cite{21}&	CSI	&CNN&	UT-HAR/Widar3&	Gesture 86.83\%, Human identification 79.61\% \\
			\hline 
			LWiHS \cite{23}&	CSI image&	ConvNeXt v2+Attention&	Widar3&	Gesture, 99.76\% across all data, without a specific scenario\\
			\hline
			PAC-CSI \cite{24}	&CSI&	Attention&	Widar3&	Gesture, IN-domain99.46\%, Corss-domain 96.47\%\\
			\hline
			WiGRUNT \cite{25}	&CSI image&	Attention&	Widar3	&Gesture, IN-domain99.71\%, Corss-domain 93.73\%\\
			\hline
			FewSense \cite{16}&	CSI&	CNN&	SignFi/Widar3	&Gesture, SignFi for traning, Widar3 for test-96.5\%\\
			\hline
			Yu Gu et al., \cite{17}&	CSI image&	Attention&	Widar3	&Gesture, IN-domain 99.69\%\\
			\hline
			Sheng Tan et al., \cite{22}&	DFS&	Spectrum Analysi&	Private	&Finger gesture 96\%\\
			\hline
			Hua Kang et al., \cite{26}&	DFS&	GRU+Attention&	Widar3&	Gesture, Corss-domain87.8\%\\
			\hline
			WiFi2Radar \cite{27}&	DFS&	DANN &	Private	&Human activities, Cross-orientations   91\%\\
			\hline
	\end{tabular}}
\end{table*}

\subsection{Motion-based features} 
Widar3 \cite{15} introduced a domain-independent feature called the Body-coordinate velocity profile (BVP), which show the distribution of signal power over velocity components across different body parts during gesture movements. WiHF \cite{20} proposed a cross-domain motion variation pattern by extracting rhythmic acceleration, deceleration, and even pauses in gesture movements of individual users, showcasing the uniqueness of gestures and users. These studies have pioneered novel ideas and methods, but involve complex computational steps and typically require multiple distributed receivers to enhance spatial resolution in order to ensure recognition accuracy. Additionally, the network proposed by the aforementioned research struggles to fully capture deep-level features information of gestures in the spatiotemporal dimensions of CSI.

\subsection{CSI-based features} 
AutoFi \cite{21} directly used sampled  CSI data for training, enabling an automatic sampling and learning WiFi sensing system without the need for data preprocessing. LWiHS \cite{23} transformed CSI amplitude into three types of images and fused them as input, which can achieve high sensing performance in complex environments with low computational complexity. The authors of \cite{24} used CSI phase as input and significantly improved in-domain and cross-domain accuracy by employing data augmentation to expand the training dataset. WiGRUNT \cite{25} transformed CSI phase into images as input and utilized computer vision techniques for recognition, achieving high in-domain results on the Widar dataset. FewSense \cite{16} used both CSI amplitude and phase as input, achieving excellent in-domain performance with only a small number of samples. The authors of \cite{17} transformed both CSI amplitude and phase into images, used self-attention to extract features from each image type, and combined them as input, further enhancing cross-orientation and cross-location recognition accuracy while maintaining high in-domain accuracy. However, the raw CSI data primarily reflects channel response information and cannot clearly capture the variations caused by gesture movements, which necessitates the network to indirectly infer gestures through changes in amplitude and phase..

\subsection{DFS-based features} 
DFS can be viewed as a refined feature derived from raw CSI data, with clear physical meaning directly related to gesture movements. It filters out a significant amount of gesture-irrelevant channel information, thereby offering benefits such as anti-interference, low dimensionality, and high efficiency. The authors of \cite{22} distinguished finger gestures by extracting DFS information and analyzing the Doppler spectrum. The authors of \cite{26} directly input DFS into DL networks, leveraging the temporal information contained within to recover gestures. WiFi2Radar \cite{27} effectively reduced dependency on multiple WiFi receivers and target domain data availability by using DL networks to repair the extracted DFS. Compared to raw CSI, DFS more accurately reflects velocity changes caused by hand movements during gestures. Therefore, we preprocess raw CSI data using conjugate elimination method, followed by STFT to extract DFS information, which is then transformed into image form. By concatenating the DFS images from each receiver as model input and building attention mechanism-based network architecture, we significantly enhance cross-domain gesture recognition capability while maintaining high in-domain gesture recognition accuracy.

\section{System Design}
In this part, we first preprocess the CSI data, and then introduce the overview of the architecture of attention-based network. Specifically, the CSI preprocessing module is to extract DFS features from the raw CSI data and visualize them as time-series images, while the gesture recognition module employs SMSA and self-attention-based channel attention mechanisms to achieve precise motion capture.

\begin{figure}[h]
	\centering
	\includegraphics[width=0.9\linewidth]{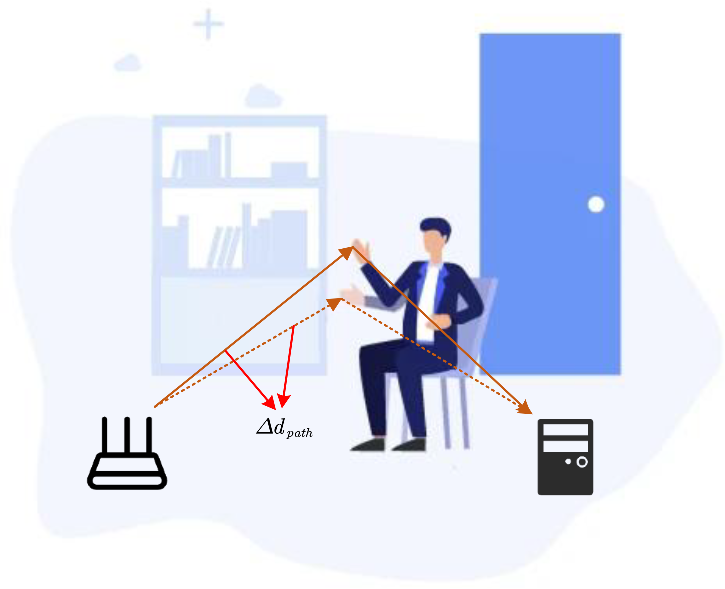}
	\caption{{WiFi Sensing Scene.} } 
	\label{fig2}
\end{figure}

\subsection{CSI Preprocessing}
As shown in Fig. \ref{fig2}, human movement causes variations in the length of the human-reflected signal path, and the rate of these path length changes can be estimated through DFS. The frequency shift of the wireless signals reflected off the moving hand is given by the following equation \cite{28}:
\begin{align}\label{dfseq}
	f_D\left( t \right) =-\frac{1}{\lambda}\frac{d}{dt}\varDelta _{\rm path},
\end{align}
where ${t}$ and ${\lambda}$ are the arrival time and wave-length, respectively, ${\varDelta _{\rm path}}$ is the change in length of the signal reflection path.

In CSI-based gesture recognition, the multipath component consists of both dynamic and static paths. Therefore, the CSI obtained at the receiver can be build as
\begin{align}
	H\text{(}f,t\text{)}=H_{\rm s}\left( f,t \right) + H_{\rm d}\left( f,t \right) ,
\end{align}
where ${f}$ is the frequency of subcarrier, ${H_{\rm s}\left( f,t \right)}$ and ${H_{\rm d}\left( f,t \right)}$ denote the static and dynamic components, respectively. It should be noted that the DFS information required for gesture recognition is contained within the dynamic component, which can be formulated as
\begin{align}
	H_{\rm d}\left( f,t \right) =\sum_{l\in P_{\rm d}}{\alpha _l\left( f,t \right) e^{j2\pi \int\limits_{-\infty}^t{f_{D_l}\left( u \right) du}}},
\end{align}
where ${\alpha _l}$ is the complex attenuation factor of the ${l}$-th path, ${P_{\rm d}}$ is the set of dynamic paths, and ${f_{D}}$ is stand for DFS shown in \eqref{dfseq}.

Since the synchronization issues and hardware impairments existed between commercial Wi-Fi transmitters and receivers, resulting in a random phase offset ${e^{-j\theta _{\rm device}}}$ in each CSI\cite{29}. The CSI obtained by Intel 5300 Wi-Fi card with random phase offset can be formulated as 
\begin{align}
\tilde{H}\text{(}f,t\text{)}=H\text{(}f,t\text{)}e^{-j\theta _{\rm device}}.
\end{align} 

As there are three antennas for Intel 5300 Wi-Fi card, and the three antennas of the same receiver exhibit identical phase offsets, we can apply conjugate multiplication between the CSI of two antennas to remove the random phase offsets. And the approach can be formulated as 
\begin{align}
	C\left( f,t \right) =&\tilde{H}_m\text{(}f,t\text{)*}\tilde{H}_{0}^{*}\text{(}f,t\text{)},
\end{align}
where ${\tilde{H}_m}$ and ${\tilde{H}_{0}}$ stand for the obtained CSI of the ${m}$-th antenna and reference antenna, respectively, and can be further expanded to \eqref{Cex} at the top of the page.

\begin{figure*}[ht]
		\begin{align}\label{Cex}
			C\left( f,t \right) &=\left( H_{s,m}\left( f,t \right) +H_{d,m}\left( f,t \right) \right) \left( H_{s,0}\left( f,t \right) +H_{d,0}\left( f,t \right) \right) ^*\notag\\
			&=\underset{1}{\underbrace{H_{s,m}\left( f,t \right) H_{s,0}^{*}\left( f,t \right) }}+\underset{2}{\underbrace{H_{s,m}\left( f,t \right) H_{d,0}^{*}\left( f,t \right) }}+\underset{3}{\underbrace{H_{d,m}\left( f,t \right) H_{s,0}^{*}\left( f,t \right) }}+\underset{4}{\underbrace{H_{d,m}\left( f,t \right) H_{d,0}^{*}\left( f,t \right) }}.
		\end{align}
	{\noindent} \rule[-10pt]{18cm}{0.05em}
\end{figure*}

In equation \eqref{Cex}, part 1 is the product of the static-path components, which can be seemed as a constant in a short time period. Part 4, which is the product of dynamic path components, has a very small value and can be ignored. Part 2 and part 3 are the products of the static paths component of one antenna and the dynamic paths component of another antenna, which both contain the DFS information. Since the DFS in part 2 is opposite to the direction we need, we increase the power of static paths component on the reference antenna by adding a value ${\beta}$, and reduce the power of the static paths component on the other antenna by subtracting a value ${\alpha}$. In this way, part 3 which contain the correct DFS information has much higher power and can be identified in the spectrogram, which shown as follow
\begin{align}
	&\bar{H}_{s,m}(f,t)H_{d,0}^{*}(f,t)=|\left[ H_{s,m}(f,t)-\alpha \right] H_{d,0}^{*}(f,t)|\notag\\
	&<|H_{d,m}(f,t)\left[ H_{s,0}^{*}(f,t)+\beta \right] |=H_{d,m}(f,t)\bar{H}_{s,0}^{*}(f,t).
\end{align}

Since part 1 can be removed via high-pass filter \cite{30} and the range of human-induced DFS is within ± 60Hz \cite{31}, we applied a 2 ${\sim}$ 60 Hz band-pass filter to denoise the data, and \eqref{Cex} can be transformed as
\begin{align}
	C\left( f,t \right)=\bar{H}_{s,m}(f,t)H_{d,0}^{*}(f,t)+H_{d,m}(f,t)\bar{H}_{s,0}^{*}(f,t).
\end{align}

Then, a principal component analysis (PCA)-based algorithm is employed to extract the principal components of the CSI streams, reducing the dimensionality of the subcarrier count. Since the carrier frequency $f$ in the above equation is a constant, the static components can be seemed as a constant in a short time period, the variable $t$ in the subsequent derivation is retained only, the Doppler spectrograms are obtained using the short-time Fourier transform (STFT) as
\begin{align}
	{\rm STFT}\{&C(\bar{f},t)\}=\bar{H}_{s,m}\int_{t-L/2}^{t+L/2}{H_{d,0}^{*}(\tau )w\left( \tau -t \right) e^{-j2\pi \bar{f}\tau}d\tau}\notag\\
	&+\bar{H}_{s,0}^{*}\int_{t-L/2}^{t+L/2}{H_{d,m}(\tau )w\left( \tau -t \right) e^{-j2\pi \bar{f}\tau}d\tau},
\end{align}
where $\bar{f}$ stand for frequency, $L$ is the window length, $w$ stand for window function, and $t$ is the center time of the window. To ensure frequency resolution, we employ gaussian window as the window function.

\begin{figure}[h]
	\centering
	\includegraphics[width=0.7\linewidth]{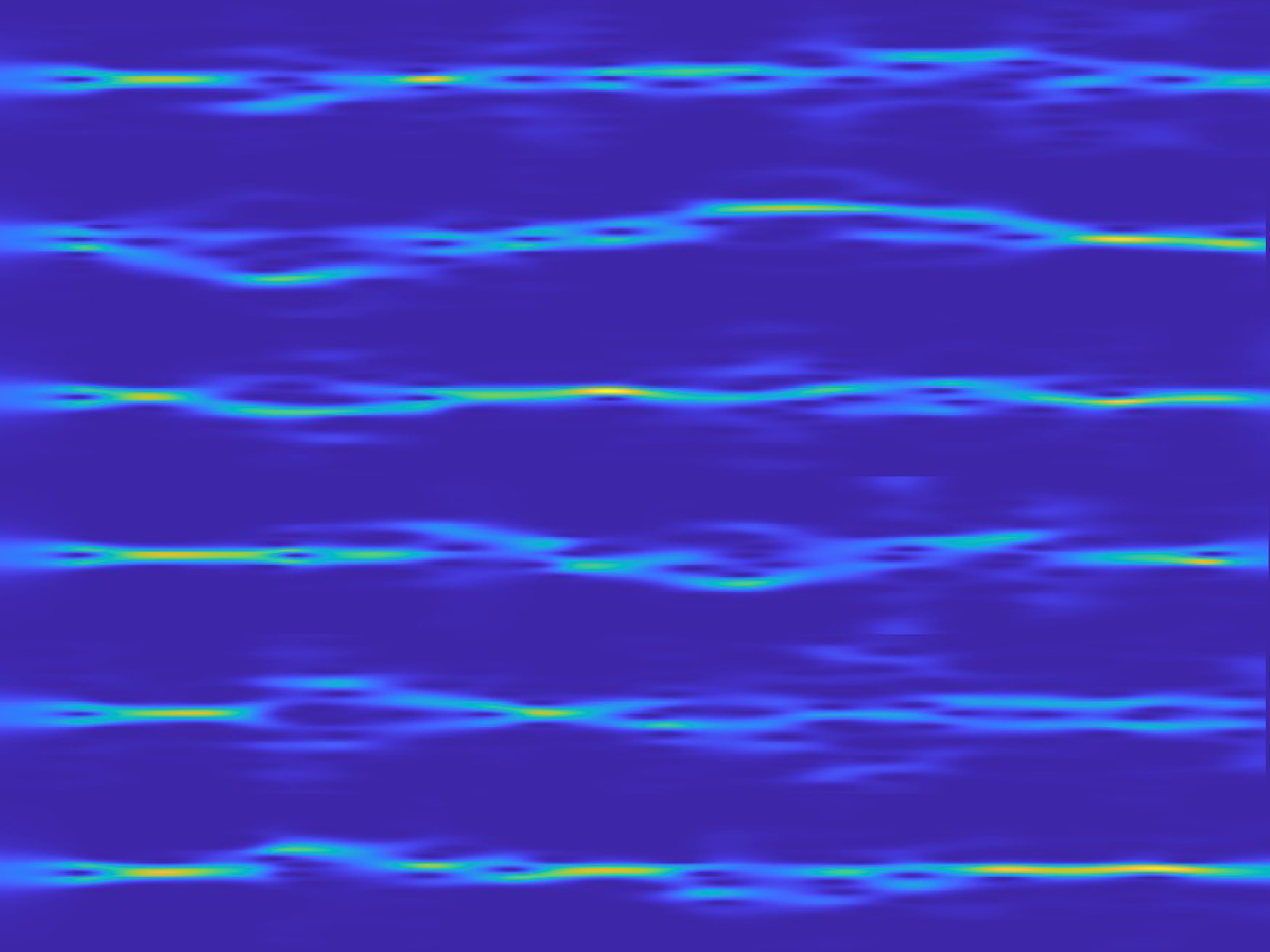}
	\caption{{The Doppler image.} } 
	\label{fig3}
\end{figure}

To transform the acquired data into a visual format, we further concatenated the Doppler spectrograms from all receivers, thereby obtaining time-series image with spatiotemporal characteristics, which is shown as Fig. \ref{fig3}. Although the extracted image still contain some prominent noise and mirrored DFS, they can already be distinguished and recognized by deep learning networks.

\subsection{Gesture Recognition}
The attention mechanism in DL is inspired by the human visual, which can help networks to focus on the most relevant historical data in the time series. By incorporating the attention mechanism, networks can automatically learn and selectively concentrate on important information in the input, thereby enhancing model performance and generalization capability.

\begin{figure*}[t]
	\centering
	\includegraphics[width=1\linewidth]{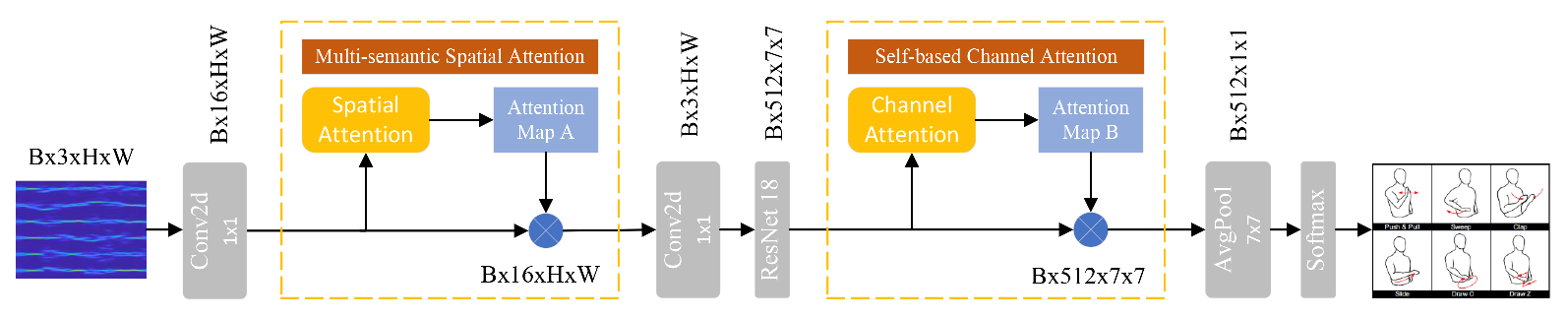}
	\caption{{The architecture of Attention-based Network.} } 
	\label{fig5}
\end{figure*}

As shown in Fig. \ref{fig5}, we integrate SMSA and self-attention-based channel attention mechanism, with ResNet18 as the backbone, aims to extract the domain-independent features of the gesture. Specifically, the preprocessed fused Doppler image undergoes channel expansion before being fed into the SMSA module to extract important features and generate attention map A. Subsequently, after combining attention map A with the input data, a 2D convolutional layer with a 1x1 kernel is used to restore the original number of data channels. Then, ResNet18 is employed to further refine features, and its output is passed into a self-attention-based channel attention module to extract important features among different channels of the feature map. This step further enhances the ability to capture domain-independent features, and generate attention map B. Finally, after combining attention map B with the output of ResNet18, a 7x7 average pooling layer is applied, and the result is given into Softmax for gesture recognition. Next, we will provide a detailed introduction to each module of the network.

\begin{figure*}[t]
	\centering
	\includegraphics[width=1\linewidth]{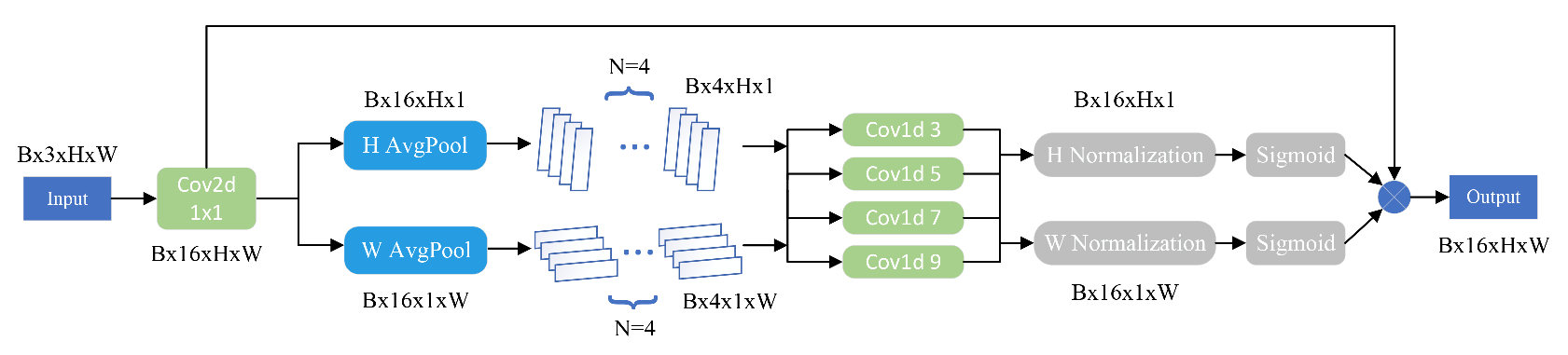}
	\caption{{The  SMSA Module.} } 
	\label{fig4}
\end{figure*}

\subsubsection{SMSA Module}
As shown in Fig. \ref{fig4} at the top of the page, the architecture of SMSA is mainly make up of four 1D convolutional layers with differing kernel sizes. To ensure that the input data can be evenly divided into four parts, a 2D convolutional layer with a kernel size of 1x1 is used to expand the number of channels in the input data from 3 to 16. Then, to capture feature information from two different perspectives, average pooling layers in the height and width directions are applied respectively, which can helps to expand the receptive field of the networks, enabling it to capture more extensive feature information. After that, the two streams of data undergone directional average pooling, are each split into four parts along the channel dimension. The sub-features can be expressed as
\begin{align}
	X_{\rm H}^{i}&=X_{\rm H}[ B,\left( i-1 \right) \times \frac{C}{4}:i\times \frac{C}{4},H,1 ], \\
	X_{\rm W}^{i}&=X_{\rm W}[ B,\left( i-1 \right) \times \frac{C}{4}:i\times \frac{C}{4},1,W ],
\end{align}
where $B$, $C$, $H$ and $W$ stand for the batch size, channel, height, and width, respectively. 
${X_{\rm H}}$ and ${X_{\rm W}}$ are the outputs of the height-direction AvgPool and width-direction AvgPool layers, respectively. The ${X^i}$ represents the $i$-th sub-feature, where $i \in [1, 4]$. 

Each part is then processed by one of four depth-shared 1D convolutional layers with kernel sizes of 3, 5, 7, and 9. The process can be expressed as
\begin{align}
	\tilde{X}_{\rm H}^{i}&=\text{Conv1d}_i\left( X_{\rm H}^{i} \right), \\
	\tilde{X}_{\rm W}^{i}&=\text{Conv1d}_i\left( X_{\rm W}^{i} \right), 
\end{align}
where $\text{Conv1d}_i$ corresponds to 1D convolutions with kernel sizes of 3, 5, 7, and 9, respectively. This parallel processing enables the extraction of multi-scale information from the data, capturing rich spatial features and enhancing the representation of both local and global characteristics.

At last, the data processed by the multi-scale 1D convolutional layers is then concatenated along the channel dimension, passed through a normalization layer and a sigmoid layer to generate two attention maps from different perspectives, which can be expressed as
\begin{align}
	{\rm Att_H}&=\sigma \left( \text{HN}\left( \text{Concat}\left( \tilde{X}_{\rm H}^{1},\tilde{X}_{\rm H}^{2},\tilde{X}_{\rm H}^{3},\tilde{X}_{\rm H}^{4} \right) \right) \right), \\
	{\rm Att_W}&=\sigma \left( \text{WN}\left( \text{Concat}\left( \tilde{X}_{\rm W}^{1},\tilde{X}_{\rm W}^{2},\tilde{X}_ {\rm W}^{3},\tilde{X}_{\rm W}^{4} \right) \right) \right), 
\end{align}
where $\sigma$ stands for the sigmoid function. $\text{HN}$ and $\text{WN}$ stand for the the height-direction normalization and width-direction normalization layers, respectively.

These attention maps are multiplied with the channel-expanded input data to produce the attention map A of the SMSA module, and can be formulated as
\begin{align}
	{\rm AttMapA}= {\rm Att_H} \times {\rm Att_W}.
\end{align}

Through this refined channel feature processing, the network can better integrate information across different feature channels, which facilitates the capture of domain-independent features and enhances the generalization capability across various scenarios of the networks.

\subsubsection{ResNet18}
The proposed network adopt 18-layer  residual networks, i.e. ResNet18, as the backbone network, and the specific structural parameters are shown in Table \ref{table9}. By incorporating residual blocks and skip connections, ResNet18 can address the issues of vanishing and exploding gradients in deep networks, allowing for virtually unlimited network depth.

\begin{table}[h]
	\centering
	\caption{ResNet18} 
	\label{table9}
	\renewcommand{\arraystretch}{2}
	\begin{tabular}{|c|c|}
		\hline  
		Module name & 18-layer \\
		\hline  
		\multirow{2}*{\makecell[c]{Processing Module}} & Conv 7 $\times$ 7, 64\\
		\cline{2-2}
		& MaxPool 3 $\times$ 3\\
		\hline 
		\multirow{4}*{\makecell[c]{Convolution Module}} & 
		$
		\left[ \begin{array}{c}
			3\times 3, 64\\
			3\times 3, 64\\
		\end{array} \right] \times 2
		$\\
		\cline{2-2}
		& $
		\left[ \begin{array}{c}
			3\times 3, 128\\
			3\times 3, 128\\
		\end{array} \right] \times 2
		$\\
		\cline{2-2}
		& $
		\left[ \begin{array}{c}
			3\times 3, 256\\
			3\times 3, 256\\
		\end{array} \right] \times 2
		$\\
		\cline{2-2}
		& $
		\left[ \begin{array}{c}
			3\times 3, 512\\
			3\times 3, 512\\
		\end{array} \right] \times 2
		$\\
		\hline
	\end{tabular}
\end{table}

\subsubsection{Self-based Channel Attention Module}
Channel attention is a mechanism that focuses on the allocation of importance across the channels of the feature map. Its primary objective is to enhance model performance by assigning distinct weights of the channels, the channels that contribute most to the task can be emphasized while suppressing the other irrelevant or redundant channels.

Different with the channel attention mechanisms that employ nested convolutional or fully-connected layers for feature extraction,  we utilize self-attention mechanisms applied to the channel dimension after compressing the spatial dimensions of the input via average and max pooling. 

\begin{figure}[h]
	\centering
	\includegraphics[width=0.7\linewidth]{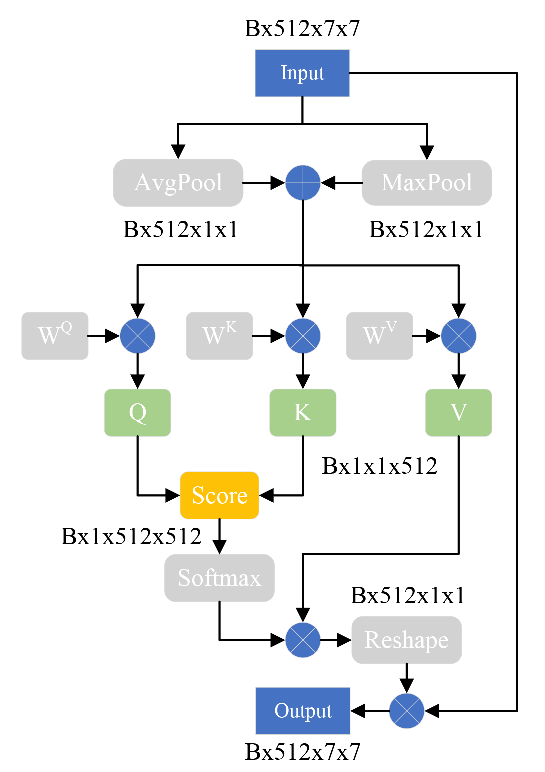}
	\caption{{The Self-based Channel Attention Module.} } 
	\label{fig6}
\end{figure}

As shown in Fig. \ref{fig6}, the ${W^Q}$, ${W^K}$ and ${W^V}$ are the weight matrices of query (Q), key (K) and value (V), respectively. The operation of the score layer can be formulated as
\begin{align}
	{\rm Score}\left( Q_i,K_j \right) =\frac{Q_i\cdot K_j}{\sqrt{d_{\rm k}}},
\end{align} 
where ${Q_i}$ is the query vector of the $i$-th channel, ${K_j}$ is the key vector of the $j$-th channel. ${\sqrt{d_{\rm k}}}$ is the scaling factor, where ${d_{\rm k}}$ stand for the dimension of the key data. After obtaining the attention scores, a softmax function is uesd to normalize the attention scores to values between 0 and 1 that sum to 1, and the attention weights can be formulated as
\begin{align}
	w_{ij}=\text{Softmax} \left( \text{Score}\left( Q_i,K_j \right) \right) .
\end{align}

By multiplying each value vector by its corresponding attention weight and summing the results, the attention map B can be obtained  as
\begin{align}
	{\rm AttMapB}=\sum_{j=1}^{512}{w_{ij}}V_j.
\end{align}

Though the self-based channel attention module, the network consider the relationship between each channel and all other channels to help  better understand the contextual information within the channels, which can enable more accurate extraction of the feature information contained inside.

\section{Experiment And Evaluation}
This section elaborates on the experimental implementation process, introduces the parameter settings, and evaluates the experimental results.

\subsection{Dataset}
\begin{figure}[h]
	\centering
	\includegraphics[width=0.8\linewidth]{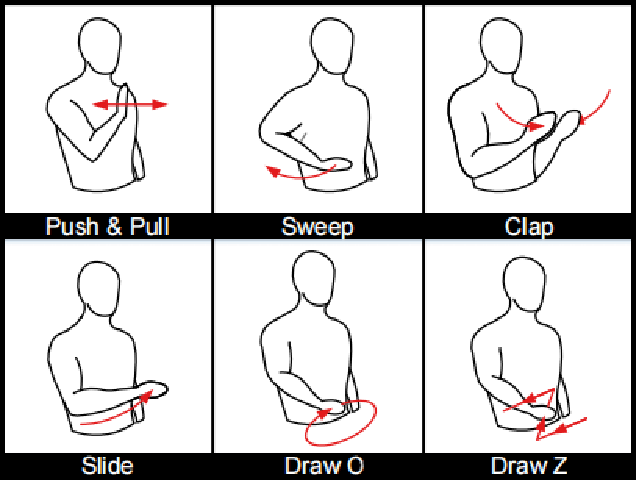}
	\caption{Types of Gestures.} 
	\label{fig7}
\end{figure}

 Widar3 is a dataset for WiFi-based gesture recognition, containing gesture data collected from diverse locations, orientations, and environments. As shown in Fig. \ref{fig7}, the dataset comprises 6 gesture categories: push $\&$ pull, sweep, clap, slide, draw-O, and draw-Z. 

\begin{table*}[ht]
	\centering
	\caption{Description Of Dataset} 
	\label{table1}
	\renewcommand{\arraystretch}{1.5}
	\begin{tabular}{|c|c|c|c|}
		\hline  
		Environments & Number of users & Gestures & Number of Samples\\
		\hline  
		Classroom & 8 & \multirow{3}*{\makecell[c]{1: Push ${\&}$ Pull; 2: Sweep; 3: Clap; 4: Slide; 5: Draw-O; 6: Draw-Z;}} & 5950\\
		\cline{1-2}
		\cline{4-4}
		Hall & 2 & & 2239\\
		\cline{1-2}
		\cline{4-4}
		Office & 4 & & 2965\\
		\hline 
	\end{tabular}
\end{table*}

For our experiments, we utilize data from 14 users across three environments in the Widar3 dataset, with each user contributing 750 samples. The original sample counts were 6,000 for the Classroom environment, 2,250 for Hall (one of users provided 1,500 samples), and 3,000 for Office. After visualizing the CSI data, we identified partial data loss in some receivers. Following the removal of incomplete data, the actual sample sizes per scenario were 5,950 for Classroom, 2,239 for Hall, and 2,965 for Office, as shown in Table \ref{table1}. 

For in-domain experiments, data from individual instances were split into training and test sets at a 9:1 ratio. For cross-domain experiments, models were trained on data from two environments and tested on the remaining third environment. Furthermore, both in-domain and cross-domain experiments provide results from 5 simulation runs, and the average accuracy serving as the performance metric.

\begin{figure}[ht]
	\centering
	\includegraphics[width=0.9\linewidth]{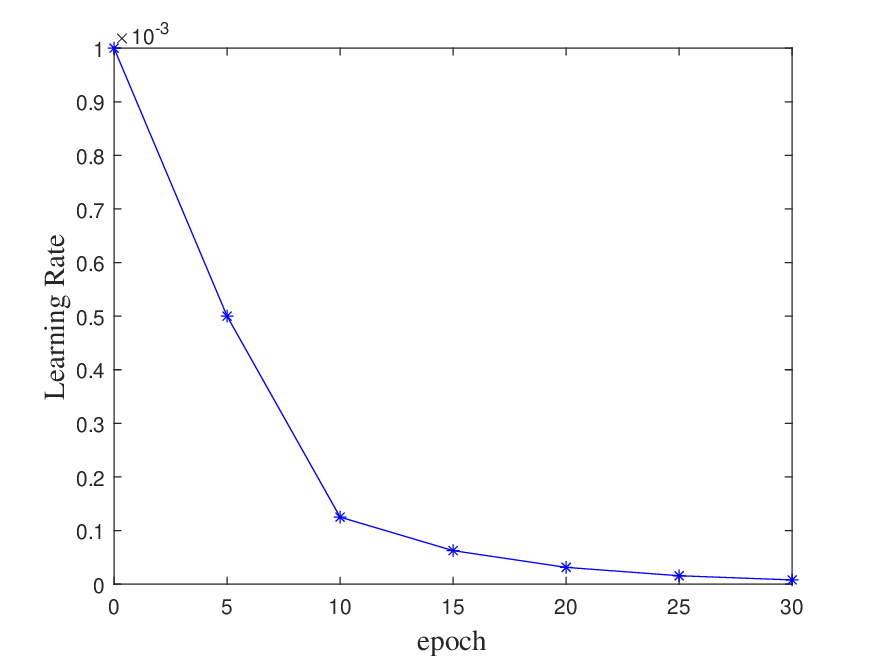}
	\caption{Learning Rate.} 
	\label{fig11}
\end{figure}
	
\begin{figure*}[ht]
	
	\begin{minipage}{0.32\linewidth}
		\centering{\includegraphics[width=\textwidth]{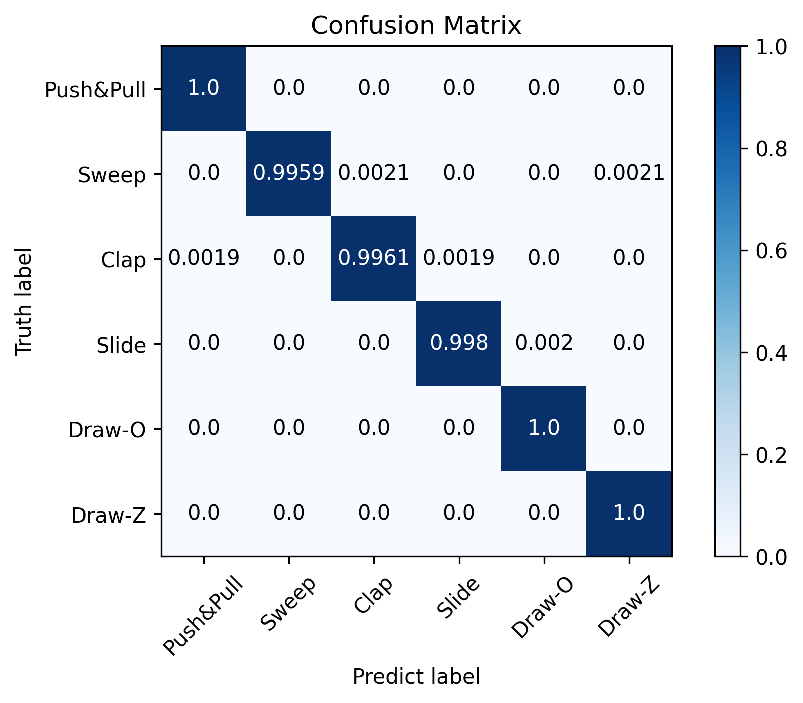}}
		\centering{a) Classroom}
	\end{minipage}
	\begin{minipage}{0.32\linewidth}
		\centering{\includegraphics[width=\textwidth]{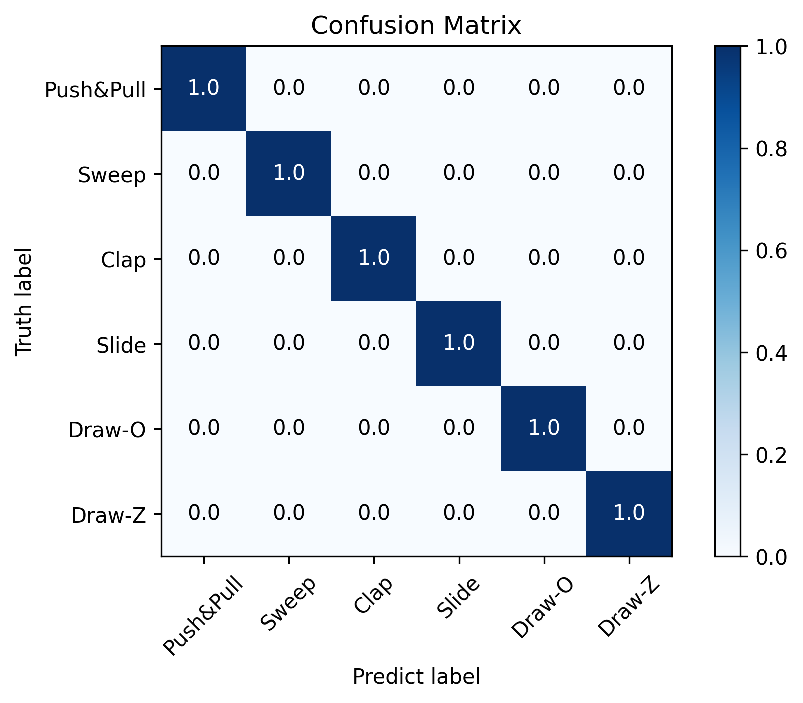}}	
		\centering{b) Hall}
	\end{minipage}
	\begin{minipage}{0.32\linewidth}
		\centering{\includegraphics[width=\textwidth]{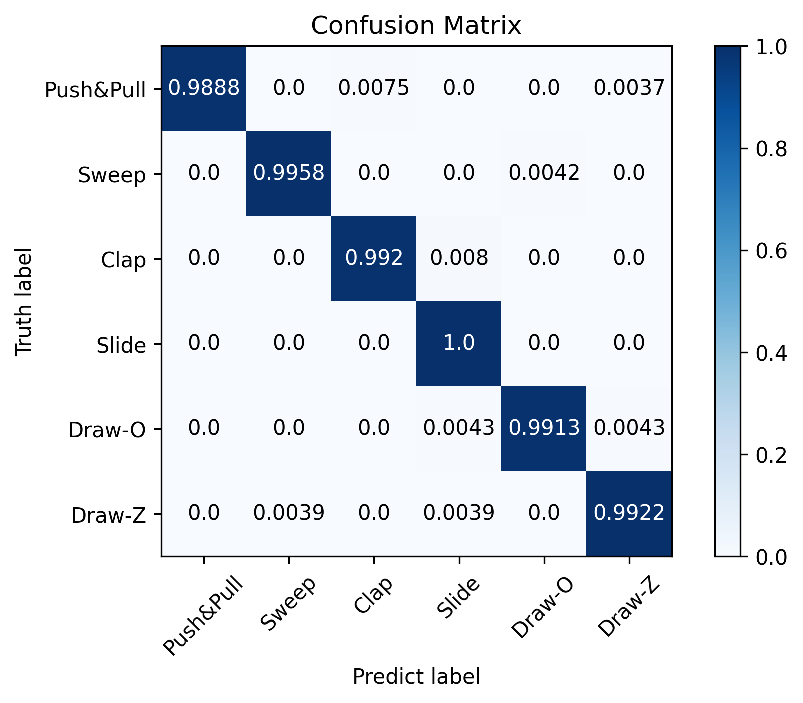}}		
		\centering{c) Office}
	\end{minipage}
	
	\caption{Confusion Matrix For In-domain Experiments.  }
	\label{fig8}
\end{figure*}
\begin{figure*}[ht]
	
	\begin{minipage}{0.32\linewidth}
		\centering{\includegraphics[width=\textwidth]{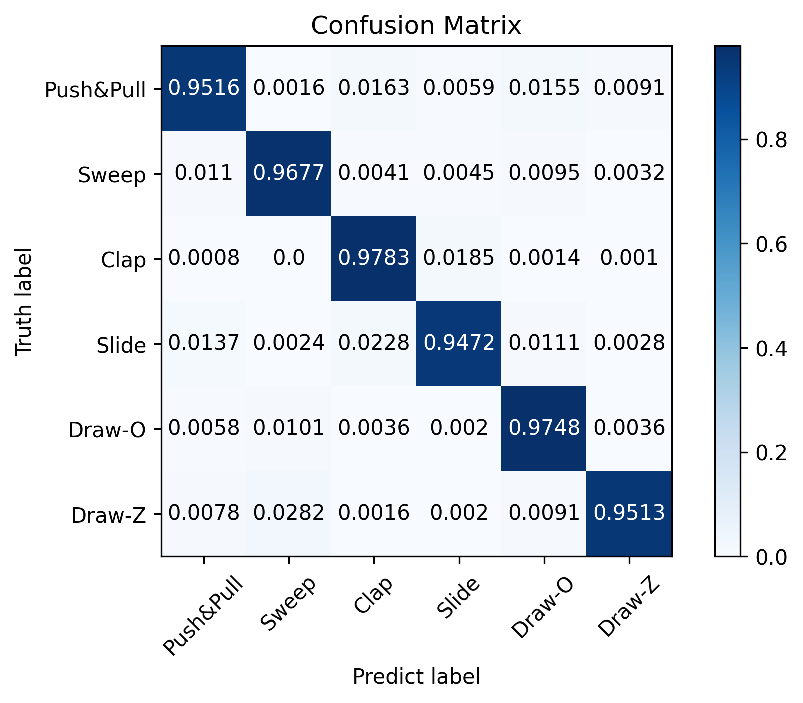}}
		\centering{a) Classroom}
	\end{minipage}
	\begin{minipage}{0.32\linewidth}
		\centering{\includegraphics[width=\textwidth]{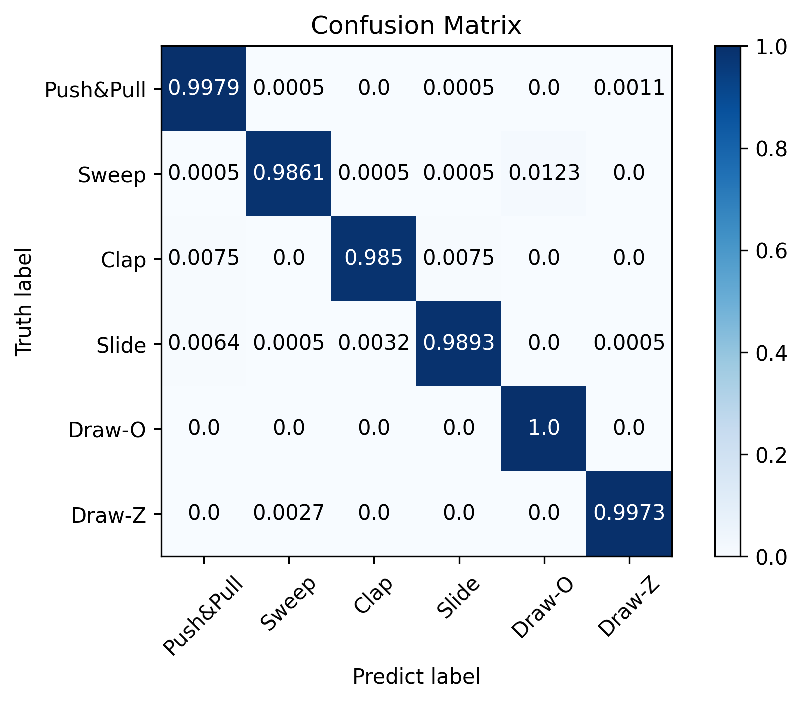}}	
		\centering{b) Hall}
	\end{minipage}
	\begin{minipage}{0.32\linewidth}
		\centering{\includegraphics[width=\textwidth]{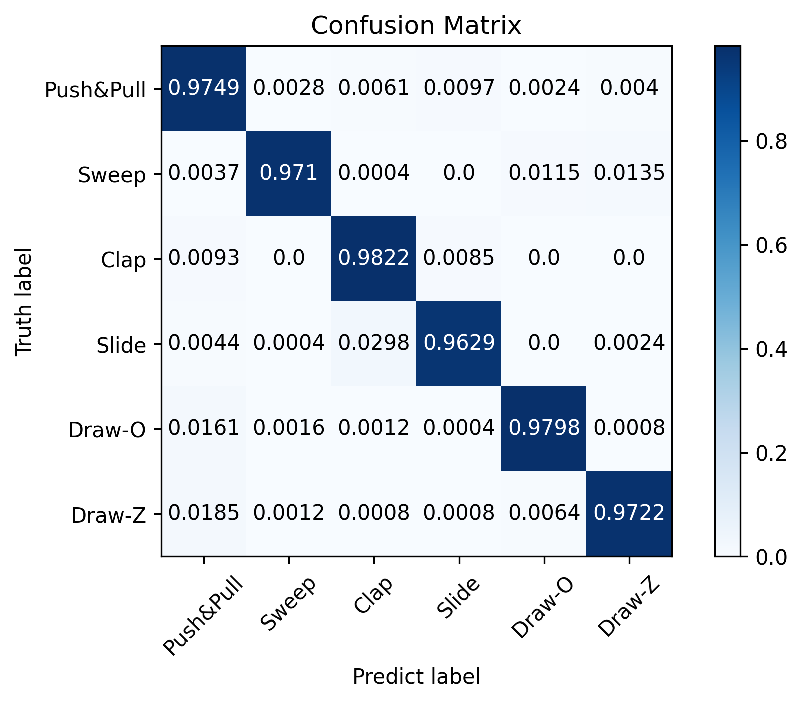}}		
		\centering{c) Office}
	\end{minipage}
	
	\caption{Confusion Matrix For Cross-domain Experiments.  }
	\label{fig9}
\end{figure*}

\subsection{HyperParameter}
The size of the epoch needs to ensure that the network converges during training, so the epoch of experiment is set to 30. Due to the limited dataset size, a small batch size of 16 was selected to ensure network performance. For the optimizer, we employed the Adam optimizer, which combines the advantages of Adagrad in handling sparse gradients and RMSprop in dealing with non-stationary objectives, resulting in faster convergence and superior performance in the network. To avoid network training falling into a local optimal value, the learning rate is set to 0.001, and the learning rate is halved every 5 training epochs, which is shown in Fig. \ref{fig11}. Given the outstanding performance of CrossEntropyLoss in classification tasks, we adopted it as the loss function to minimize the divergence between the predicted probability distribution of the network output and the true label distribution.

\subsection{In-domain Experiment}
\begin{table}[h]
	\centering
	\caption{Accuracy for In-domain Experiment} 
	\label{table2}
	\scalebox{0.9}{
	\renewcommand{\arraystretch}{1.5}
	\begin{tabular}{|c|c|c|c|c|c|c|}
		\hline  
		Environments & 1 & 2 & 3 & 4 & 5 & Avg-Acc\\
		\hline  
		Classroom & 100${\%}$ & 99.66${\%}$  & 99.83${\%}$ & 99.66${\%}$ & 100${\%}$ & \textbf{99.83${\%}$}\\
		\hline
		Hall & 100${\%}$ & 100${\%}$ & 100${\%}$ & 100${\%}$ & 100${\%}$ & \textbf{100${\%}$}\\
		\hline
		Office & 99${\%}$ & 99.32${\%}$ & 99.32${\%}$ & 99.32${\%}$ & 99.66${\%}$ & \textbf{99.32${\%}$}\\
		\hline 
	\end{tabular}}
\end{table}

As shown in Table \ref{table2}, the average accuracy rates in the Classroom, Hall, and Office environments were 99.83\%, 100\%, and 99.32\%, respectively, resulting in an overall average accuracy of 99.72\% for the in-domain experiments.

Fig. \ref{fig8} shows the confusion matrices of the in-domain experimental results. It can be observed that the performance in the Classroom and Office environments is inferior to that in the Hall environment. This is because both the Classroom and Office environments contain densely distributed objects, which generate more multipath propagation paths and consequently increase interference.

\subsection{Cross-domain Experiment}

\begin{table}[h]
	\centering
	\caption{Accuracy for Cross-domain Experiment} 
	\label{table3}
	\scalebox{0.85}{
		\renewcommand{\arraystretch}{1.5}
		\begin{tabular}{|c|c|c|c|c|c|c|}
			\hline  
			Environments & 1 & 2 & 3 & 4 & 5 & Avg-Acc\\
			\hline  
			Classroom & 96.71${\%}$ & 96.44${\%}$  & 95.83${\%}$ & 95.95${\%}$ & 95.99${\%}$ & \textbf{96.18${\%}$}\\
			\hline
			Hall & 99.33${\%}$ & 99.33${\%}$ & 99.20${\%}$ & 99.20${\%}$ & 99.24${\%}$ & \textbf{99.26${\%}$}\\
			\hline
			Office & 97.61${\%}$ & 97.44${\%}$ & 97${\%}$ & 97.47${\%}$ & 97.4${\%}$ & \textbf{97.38${\%}$}\\
			\hline 
	\end{tabular}}
\end{table}

As shown in Table \ref{table3}, the average accuracy rates in the Classroom, Hall, and Office environments were 96.18\%, 99.26\%, and 97.38\%, respectively, resulting in an overall average accuracy of 97.61\% for the cross-domain experiments.

The confusion matrices of the cross-domain experimental results is shown in Fig. \ref{fig9}. It can be seen from the confusion matrix, during cross-domain gesture recognition, the Clap and Slide gestures are prone to misclassification due to their similar hand trajectory movements. Similarly, misclassification occurs between the Sweep and Draw-O gestures because the motion trajectory of the former is contained within the latter.

\subsection{Comparative Study}
 \begin{table}[h]
	\centering
	\caption{Overall Accuracy Compare To Exist Networks} 
	\label{table4}
	\scalebox{1}{
		\renewcommand{\arraystretch}{1.5}
		\begin{tabular}{|c|c|c|}
			\hline  
			Networks & In-domain & Cross-domain \\
			\hline  
			Widar3 & 92.7${\%}$ & 92.4${\%}$  \\
			\hline
			WiHF & 97.65${\%}$ & 91.07${\%}$ \\
			\hline
			WiGRUNT & 99.71${\%}$ & 93.73${\%}$ \\
			\hline 
			PAC-CSI & 99.46${\%}$ & 96.47${\%}$ \\
			\hline
			\textbf{Our} & \textbf{99.72${\%}$} & \textbf{97.61${\%}$} \\
			\hline
	\end{tabular}}
\end{table}

\begin{table}[h]
	\centering
	\caption{Detailed Accuracy Comparison For Cross-domain} 
	\label{table5}
	\scalebox{1}{
		\renewcommand{\arraystretch}{1.5}
		\begin{tabular}{|c|c|c|c|}
			\hline  
			Networks & Cross-Classroom & Cross-Hall & Cross-Office\\
			\hline  
			WiGRUNT & 87.9${\%}$ & 97.82${\%}$ & 95.47${\%}$\\
			\hline 
			\textbf{Our} & \textbf{96.18${\%}$} & \textbf{99.26${\%}$} & \textbf{97.38${\%}$}\\
			\hline
	\end{tabular}}
\end{table}

\begin{figure}[h]
	\centering
	\includegraphics[width=1\linewidth]{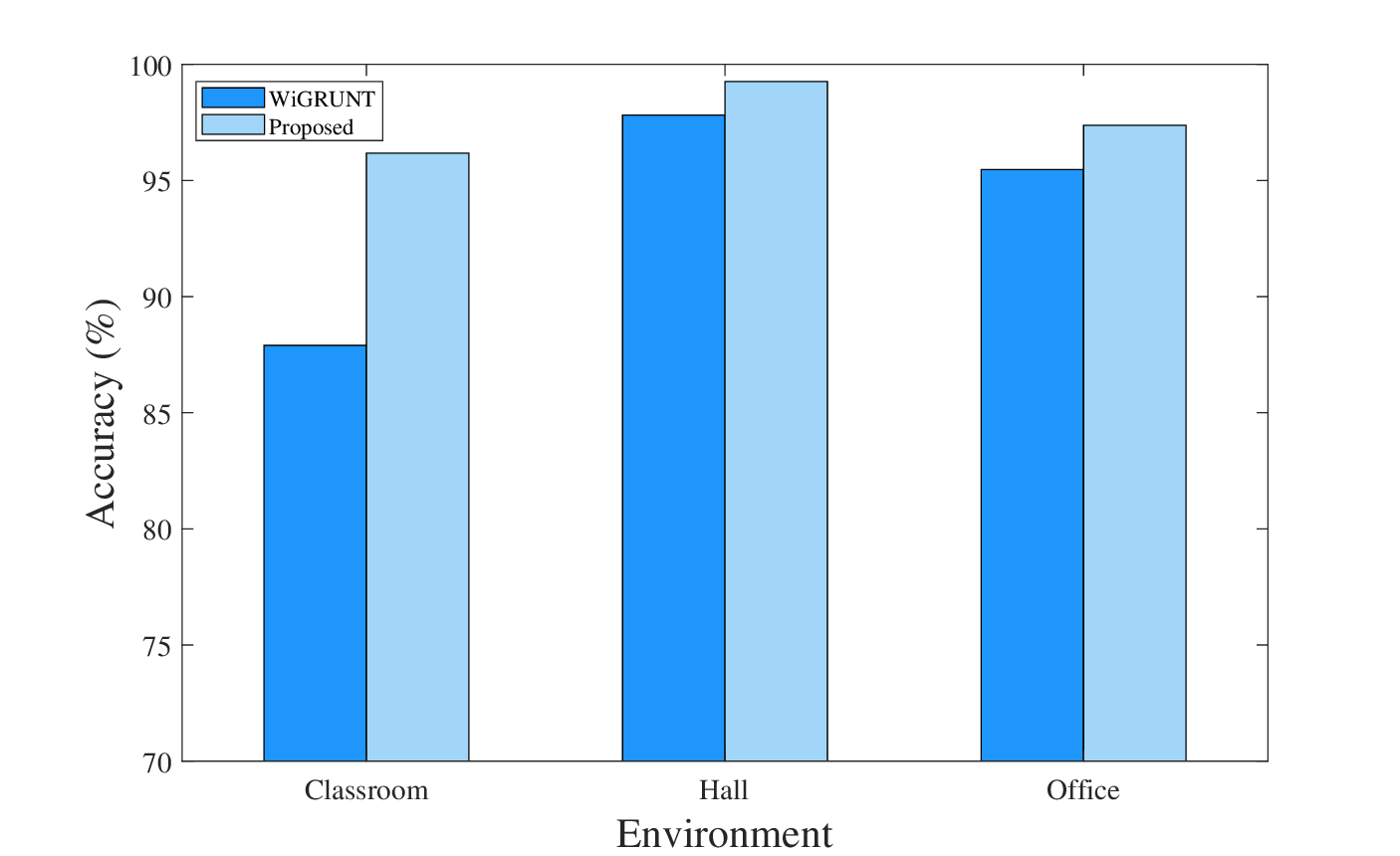}
	\caption{Detailed Accuracy Comparison.} 
	\label{fig10}
\end{figure}

 In this part, we conducted comparisons with the classical networks Widar3 \cite{15} and WiHF \cite{20}, as well as the state-of-the-art networks WiGRUNT \cite{25} and PAC-CSI \cite{24}, which have demonstrated exceptional performance, and all experimental results are derived from the Widar3 dataset. As shown in Table \ref{table4}, the proposed network has achieved the highest average in-domain accuracy of 99.72\% among the currently top-performing networks, demonstrating a significant improvement over the Widar3 network while maintaining comparable in-domain performance with the WiGRUNT network. While maintaining in-domain performance, the proposed network achieves further improvement in cross-domain performance, even surpassing WiHF's in-domain accuracy, and demonstrates a 1.14\% enhancement compared to the PAC-CSI network.

Since the PAC-CSI network, which demonstrated outstanding performance in the comparative study, did not provide specific cross-domain experimental data, we conducted a detailed comparison with the WiGRUNT network. The WiGRUNT network exhibits a 3.74\% lower average cross-domain accuracy compared to the proposed network. As shown in Table \ref{table5}, the proposed network achieves 1.44\% and 1.91\% higher cross-domain performance than WiGRUNT in the Hall and Office environments, respectively. However, WiGRUNT's cross-domain performance in the Classroom environment is notably poor at only 87.9\%. As shown in Fig. \ref{fig10}, the proposed network elevates the cross-domain performance in the Classroom environment to a comparable level with the other two environments, reaching 96.18\% with an 8.28\% improvement over WiGRUNT, which shown the significant generalization capability of our network.

\subsection{Impact of Split Proportion}

\begin{figure}[h]
	\centering
	\includegraphics[width=1\linewidth]{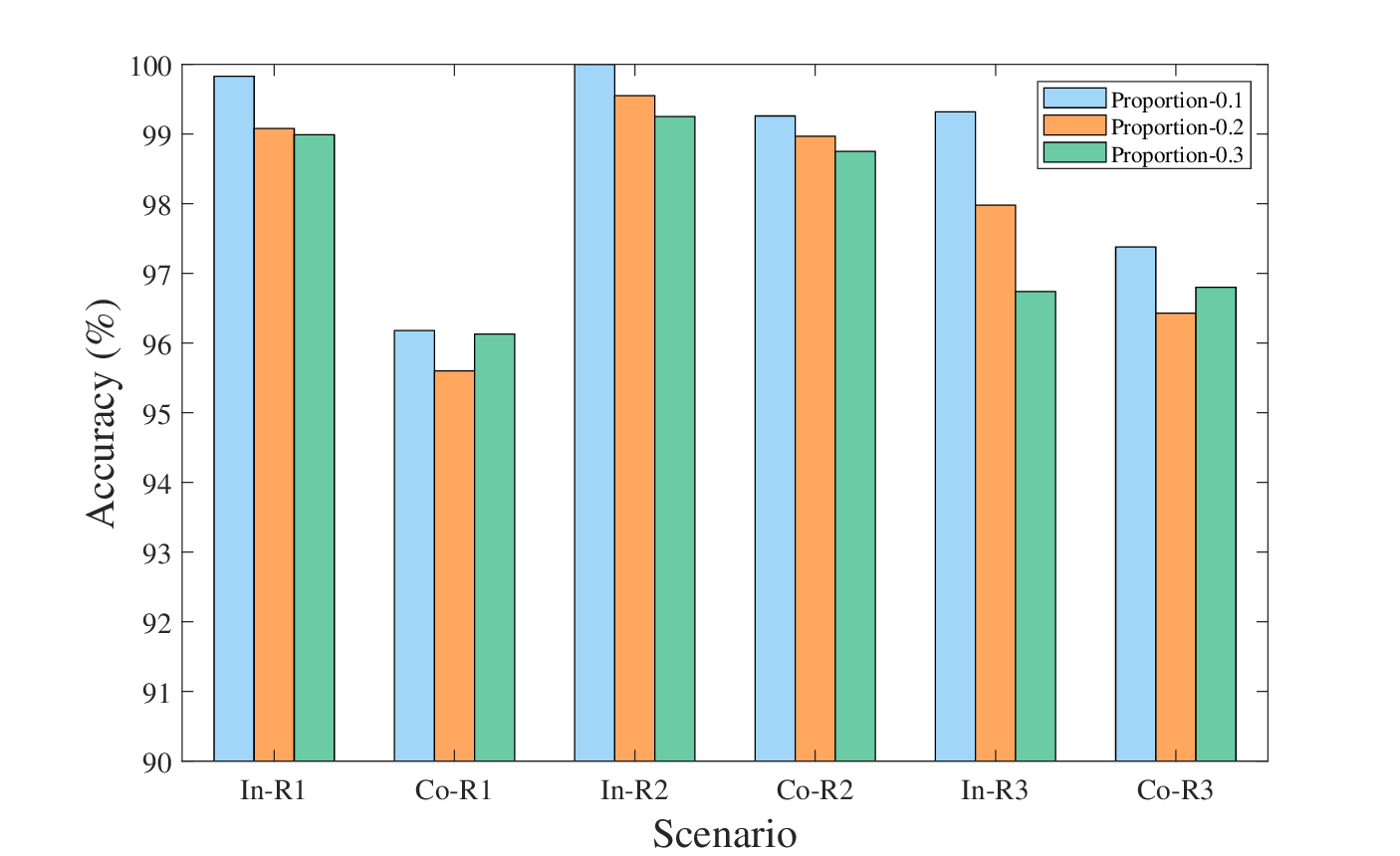}
	\caption{Impact of Proportion.} 
	\label{fig14}
\end{figure}

To evaluate the network's sensitivity to the volume of training data, we conducted both in-domain and cross-domain experiments with varying proportion of training to testing data. Specifically, we trained the network using dataset split proportion of 0.1, 0.2, and 0.3, where larger split proportion correspond to smaller amounts of training data. As shown in Fig. \ref{fig14}, the in-domain experiments were primarily affected by the split proportion, with accuracy differences exceeding 1\% between proportion of 0.1 and 0.3. In contrast, cross-domain accuracy fluctuated within 0.5\%. This is because the cross-domain experiments utilized the combined data from two environments, which is substantially larger than the data from any single environment. However, as the split proportion increases, the impact on cross-domain accuracy becomes more pronounced. These results indicate that while the proposed network achieves superior performance, it requires a substantial amount of training data, and insufficient data volume fails to fully realize its potential.

\subsection{Impact of Location and Orientation}

\begin{figure*}[ht]
	
	\begin{minipage}{0.32\linewidth}
		\centering{\includegraphics[width=\textwidth]{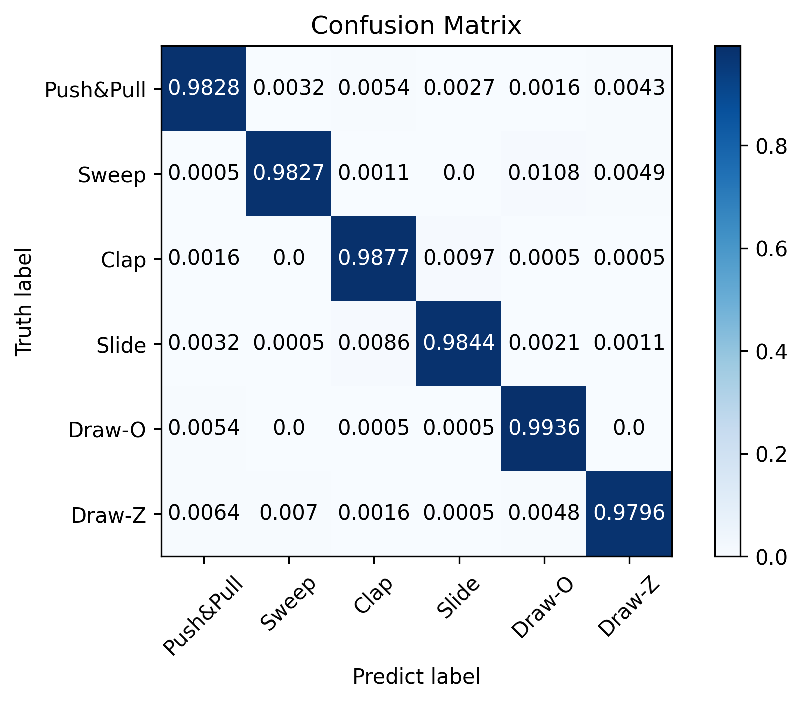}}
		\centering{a) Location}
	\end{minipage}
	\begin{minipage}{0.32\linewidth}
		\centering{\includegraphics[width=\textwidth]{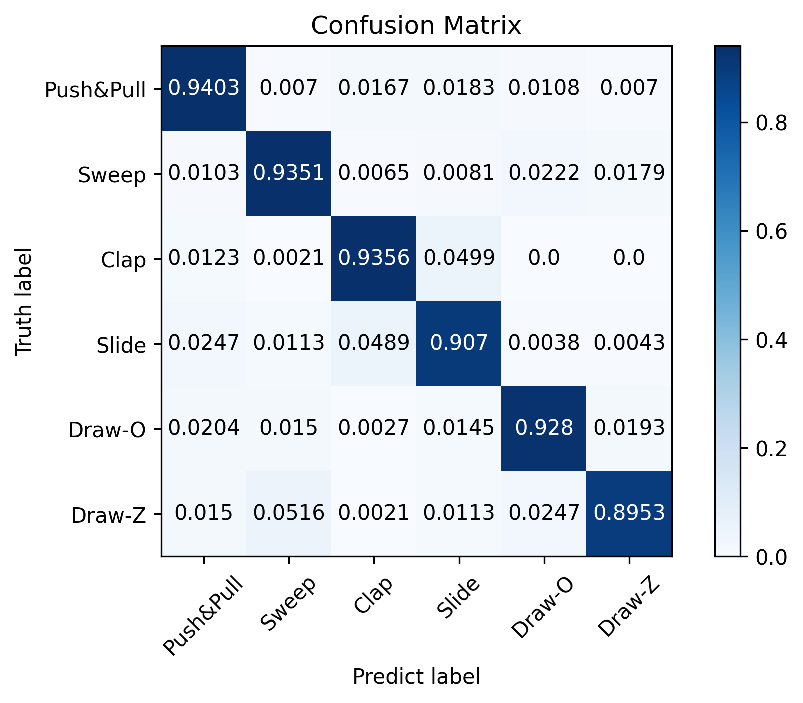}}	
		\centering{b) Orientation}
	\end{minipage}
		\begin{minipage}{0.32\linewidth}
		\centering{\includegraphics[width=\textwidth]{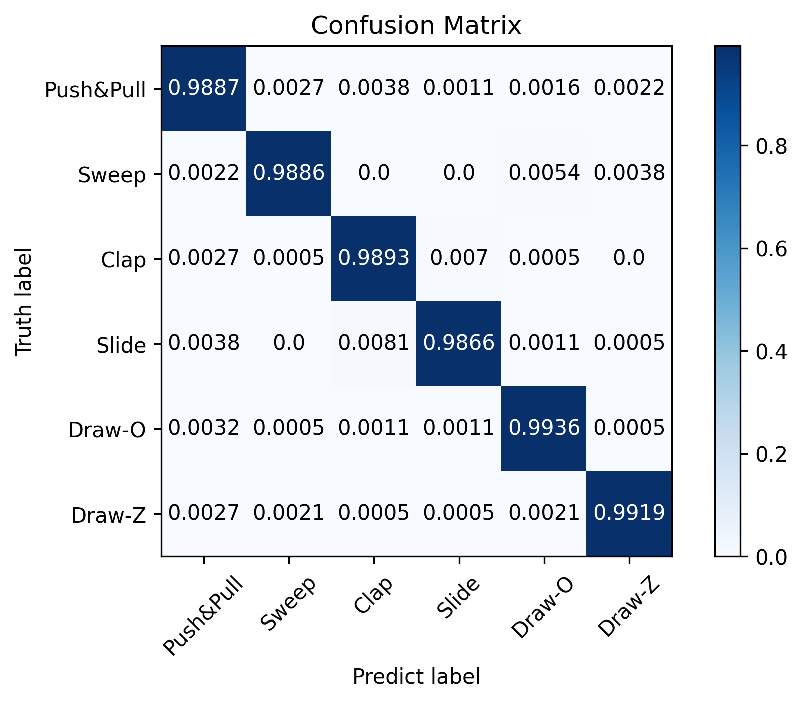}}	
		\centering{c) User}
	\end{minipage}
	
	\caption{Confusion Matrix For Impact Experiments.  }
	\label{fig15}
\end{figure*}

\begin{figure}[h]
	\centering
	\includegraphics[width=0.8\linewidth]{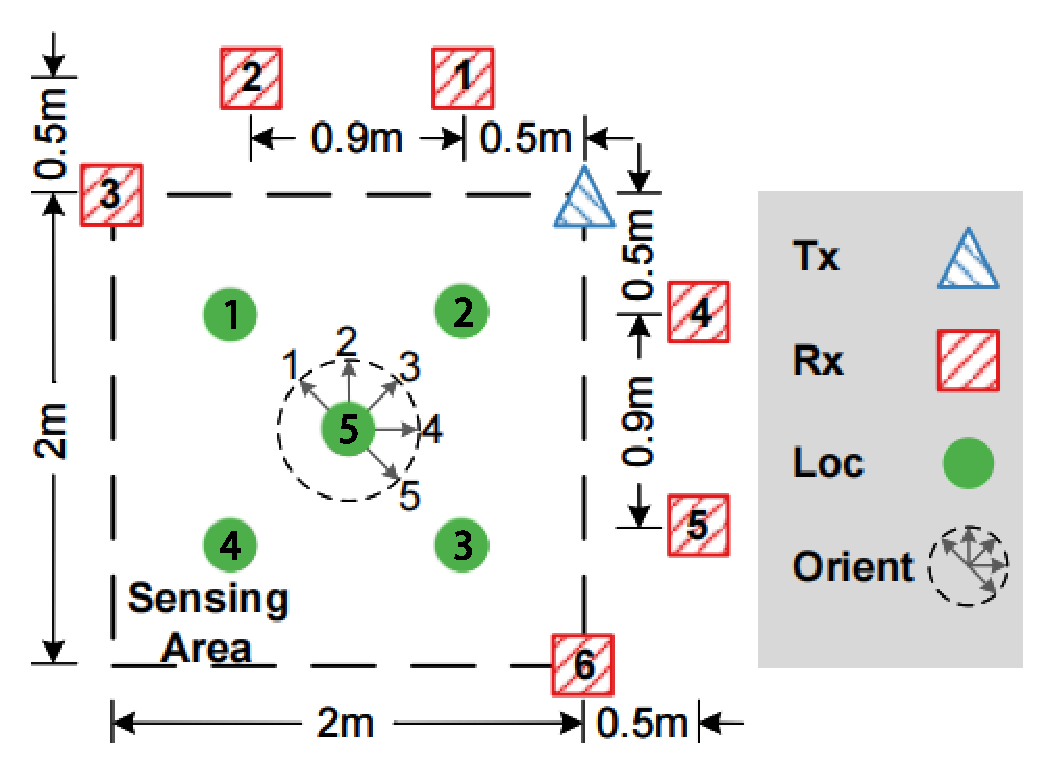}
	\caption{Layout of Widar3.} 
	\label{fig13}
\end{figure}

Different locations and orientations may cause occlusion of some receivers during data collection. To investigate the impact of varying collection locations and orientations on network training, we conducted cross-location and cross-orientation experiments. As shown in the Fig. \ref{fig15}, the accuracy rates for each gesture in both cross-location and cross-orientation experiments remain at consistent levels, with average accuracy rates of 98.51\% and 92.36\%, respectively. The gesture data of Widar3 dataset have 5 locations and 5 orientations, which is shown in Fig. \ref{fig13}. In this case, we integrated the data from all three environments and performed classification based on orientation labels and location labels separately, to investigate the impact of data collection orientation and location on network training.

\begin{table}[h]
	\centering
	\caption{Accuracy for Cross-location/orientation Experiment} 
	\label{table7}
	\scalebox{0.9}{
		\renewcommand{\arraystretch}{1.5}
		\begin{tabular}{|c|c|c|c|c|c|}
			\hline  
			Scenario & 1 & 2 & 3 & 4 & 5 \\
			\hline  
			WiGRUNT-Location & 97.33${\%}$ & 93${\%}$  & 96.56${\%}$ & 97.67${\%}$ & 98.56${\%}$ \\
			\hline
			\textbf{Our-Location} & \textbf{98.07${\%}$} & \textbf{97.85${\%}$}  & \textbf{98.74${\%}$} & \textbf{98.48${\%}$} & \textbf{99.42${\%}$} \\
			\hline
			WiGRUNT-Orientation & 90.89${\%}$ & 95.67${\%}$  & 92.56${\%}$ & 96.89${\%}$ & 93.22${\%}$ \\
			\hline
			\textbf{Our-Orientation} & \textbf{83.52${\%}$} & \textbf{95.93${\%}$}  & \textbf{97.23${\%}$} & \textbf{98.07${\%}$} & \textbf{87.04${\%}$} \\
			\hline 
	\end{tabular}}
\end{table}

As shown in Table \ref{table7}, in the cross-location experiments, the proposed network achieved consistently similar performance levels across different locations, demonstrating that variations in data collection locations had minimal impact on its performance. However, in the cross-orientation experiments, a significant decline in accuracy was observed for orientation 1 and orientation 5, indicating that data collection orientation substantially affects network performance. To explain this phenomenon, we refer to Fig. \ref{fig13}. When data is collected facing orientation 1, receivers 4, 5, and 6 are entirely behind the human body, whereas when facing orientation 5, receivers 1, 2, and 3 are completely obscured by the human body. It is important to note that gesture actions primarily involve arm movements in front of the body. Consequently, WiFi signals received by the obstructed receivers mainly reflect off the static human torso, rather than capturing dynamic gesture information. As a result, the fused DFS images fail to provide sufficient discriminative features for the network to learn gesture characteristics effectively, thereby impairing training performance and leading to poor recognition accuracy in these two orientations.

To illustrate the effectiveness of proposed network, we conducted a detailed comparison with the WiGRUNT network. As shown in Table \ref{table7}, in the cross-location experiments, the proposed network achieved an improvement of approximately 1\% at each location, with a notable enhancement of 3.15\% at location 2. For the cross-orientation experiments, while the network's performance was more susceptible to data collection orientations, the proposed network exhibits greater sensitivity and performs worse than WiGRUNT in orientation 1 and orientation 5, where WiGRUNT itself already demonstrates poor performance, but it still demonstrated improvements ranging from 0.3\% to 4.5\% in non-extreme orientation scenarios.

\subsection{Impact of User}

\begin{table}[h]
	\centering
	\caption{Accuracy for Cross-User} 
	\label{table8}
	\scalebox{0.9}{
		\renewcommand{\arraystretch}{1.5}
		\begin{tabular}{|c|c|c|c|c|c|c|}
			\hline  
			user1 & user3 & user5 & user6 & user7 & user8 & user9 \\
			\hline  
			\textbf{98.8 ${\%}$} & \textbf{99.73${\%}$} & \textbf{98.53${\%}$}  & \textbf{99.6${\%}$} & \textbf{97.45${\%}$} & \textbf{96.5${\%}$} & \textbf{99.18${\%}$}\\
			\hline
            user10 & user11 & user12 & user13 & user14 & user15 & user16 \\
			\hline
			\textbf{98.65${\%}$} & \textbf{98.52${\%}$} & \textbf{99.06${\%}$}  & \textbf{99.33${\%}$} & \textbf{99.73${\%}$} & \textbf{100${\%}$} & \textbf{100${\%}$}\\
			\hline 
	\end{tabular}}
\end{table}

Individuals have different body shapes and sizes, and gesture actions include behavioral individuality. To investigate the impact of data collected from different individuals on network training, we conducted tests and comparisons on 14 users from the dataset used. In Table \ref{table8}, each user ID corresponds to a user ID in the Widar3 dataset. The network was trained using data from all other users and tested on the corresponding user. As shown in Table \ref{table8} and Fig. \ref{fig15}, the average accuracy of the cross-user experiments reached 98.92\%, and the maximum downward deviation of individual user accuracy from the average was 2.42\%. This indicates that data collected from different individuals has minimal impact on the network, further demonstrating the strong generalization capability of the proposed network.

\subsection{Hardware Platform}
\begin{figure}[h]
	\centering
	\includegraphics[width=0.8\linewidth]{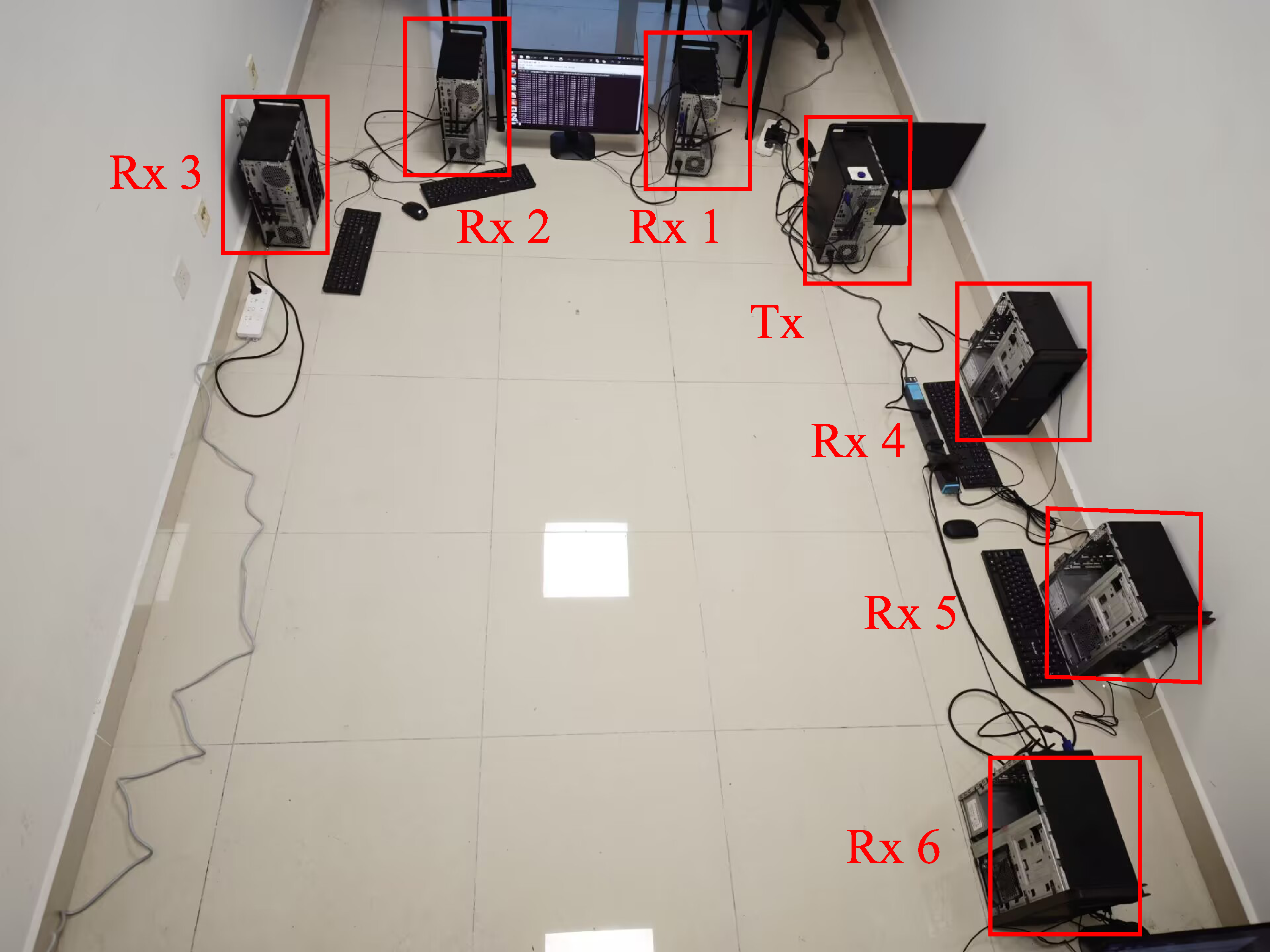}
	\caption{Hardware Platform.} 
	\label{fig12}
\end{figure}

To further advance research on WiFi sensing technology, our team has established a WiFi sensing hardware platform using seven host machines (one transmitter and six receivers), which is shown in Fig. \ref{fig12}. All transceivers are commercially available desktop computers equipped with Intel 5300 wireless network cards. The Linux CSI tool \cite{32} was installed on each device for logging CSI measurements, along with the Ubuntu operating system, and all units were connected to the campus public WiFi for debugging. Regarding the scenario layout, we followed the hardware configuration of Widar3 as illustrated in Fig. \ref{fig13}. This hardware platform has been fully debugged and is capable of collecting CSI data.

\section{Conclusion}
Within the field of ISAC, we have designed a gesture recognition system based on WiFi sensing technology that demonstrates significant cross-domain performance. This system employs a conjugate multiplication method to effectively address the phase offset present in the CSI received by the receivers. Subsequently, we extracted DFS information from the processed CSI data and converted it into fused images for training visual network. The network architecture incorporated SMSA mechanisms and self-attention-based channel attention mechanisms, enabling high-accuracy gesture recognition. In experiments conducted on the Widar3 dataset, the system achieved a high average in-domain accuracy of 99.72\% and attained a remarkable cross-domain accuracy of 97.61\%. Furthermore, during cross-domain testing on the Widar3 dataset, our system consistently achieved high accuracy levels across all three environments, fully demonstrating its generalization capability. These results validated significant breakthroughs compared to existing methods and confirm the system's high performance and practical utility. While this study is based on a manually segmented gesture dataset focusing on single-user scenarios, multi-user scenarios are inevitable in practical gesture recognition applications. Therefore, we have established a WiFi sensing hardware platform to further explore related technologies and applications, aiming to develop a more comprehensive and multimodal integrated ISAC system in future research.

	\bibliographystyle{IEEEtran}
	\bibliography{IEEEabrv,Refer}
	
\end{document}